\documentclass[sigconf]{acmart}
\AtBeginDocument{%
  \providecommand\BibTeX{{%
    \normalfont B\kern-0.5em{\scshape i\kern-0.25em b}\kern-0.8em\TeX}}}

\copyrightyear{2024}
\acmYear{2024}
\setcopyright{acmlicensed}\acmConference[HRI '24 Companion]{Companion of the 2024 ACM/IEEE International Conference on Human-Robot Interaction}{March 11--14, 2024}{Boulder, CO, USA}
\acmBooktitle{Companion of the 2024 ACM/IEEE International Conference on Human-Robot Interaction (HRI '24 Companion), March 11--14, 2024, Boulder, CO, USA}
\acmDOI{XX.XXXX/XXXXXXX.XXXXXXX}
\acmISBN{XXX-X-XXXX-XXXX-X/XX/XX}

\begin{document}

\title{CognitiveDog: Large Multimodal Model Based System to Translate Vision and Language into Action of Quadruped Robot}


\author{Artem Lykov}
\email{artem.lykov@skoltech.ru}
\orcid{0000-0001-6119-2366} 
\affiliation{%
      \institution{Skolkovo Institute of Science and Technology}
      \city{} 
      \country{}
}

\author{Mikhail Litvinov}
\email{mikhail.litvinov2@skoltech.ru}
\orcid{0009-0009-0423-0499}
\affiliation{%
      \institution{Skolkovo Institute of Science and Technology}
      \city{} 
      \country{}
}

\author{Mikhail Konenkov}
\email{mikhail.konenkov@skoltech.ru}
\orcid{0009-0003-5979-487X}
\affiliation{%
      \institution{Skolkovo Institute of Science and Technology}
      \city{} 
      \country{}
}

\author{Rinat Prochii}
\email{rinat.prochii@skoltech.ru}
\orcid{0009-0008-0816-7442}
\affiliation{%
      \institution{Skolkovo Institute of Science and Technology}
      \city{} 
      \country{}
}

\author{Nikita Burtsev}
\email{nikita.burtsev@skoltech.ru}
\orcid{0009-0006-7297-6901}
\affiliation{
      \institution{Skolkovo Institute of Science and Technology}
      \city{} 
      \country{}
}

\author{Ali Alridha Abdulkarim}
\email{ali.abdulkarim@skoltech.ru}
\orcid{0009-0008-0613-5359}
\affiliation{
      \institution{Skolkovo Institute of Science and Technology}
      \city{} 
      \country{}
}

\author{Artem Bazhenov}
\email{artem.bazhenov@skoltech.ru}
\orcid{0009-0001-2228-7298}
\affiliation{
      \institution{Skolkovo Institute of Science and Technology}
      \city{} 
      \country{}
}

\author{Vladimir Berman}
\email{vladimir.berman@skoltech.ru}
\orcid{0000-0003-4530-3518}
\affiliation{
      \institution{Skolkovo Institute of Science and Technology}
      \city{} 
      \country{}
}

\author{Dzmitry Tsetserukou}
\orcid{0000-0001-8055-5345}
\email{d.tsetserukou@skoltech.ru}
\affiliation{
      \institution{Skolkovo Institute of Science and Technology}
      \city{} 
      \country{}
}

\renewcommand{\shortauthors}{Lykov et al.}

\renewcommand{\shorttitle}{CognitiveDog}

\begin{abstract}
This paper introduces CognitiveDog, a pioneering development of quadruped robot with Large Multi-modal Model (LMM) that is capable of not only communicating with humans verbally but also physically interacting with the environment through object manipulation. The system was realized on Unitree Go1 robot-dog equipped with a custom gripper and demonstrated autonomous decision-making capabilities, independently determining the most appropriate actions and interactions with various objects to fulfill user-defined tasks. These tasks do not necessarily include direct instructions, challenging the robot to comprehend and execute them based on natural language input and environmental cues. The paper delves into the intricacies of this system, dataset characteristics, and the software architecture. Key to this development is the robot's proficiency in navigating space using Visual-SLAM, effectively manipulating and transporting objects, and providing insightful natural language commentary during task execution. 
Experimental results highlight the robot's advanced task comprehension and adaptability, underscoring its potential in real-world applications. The dataset used to fine-tune the robot-dog behavior generation model is provided at the following link: \textbf{huggingface.co/datasets/ArtemLykov/CognitiveDog\_dataset}
\end{abstract}



\begin{teaserfigure}
  \includegraphics[width=\textwidth]{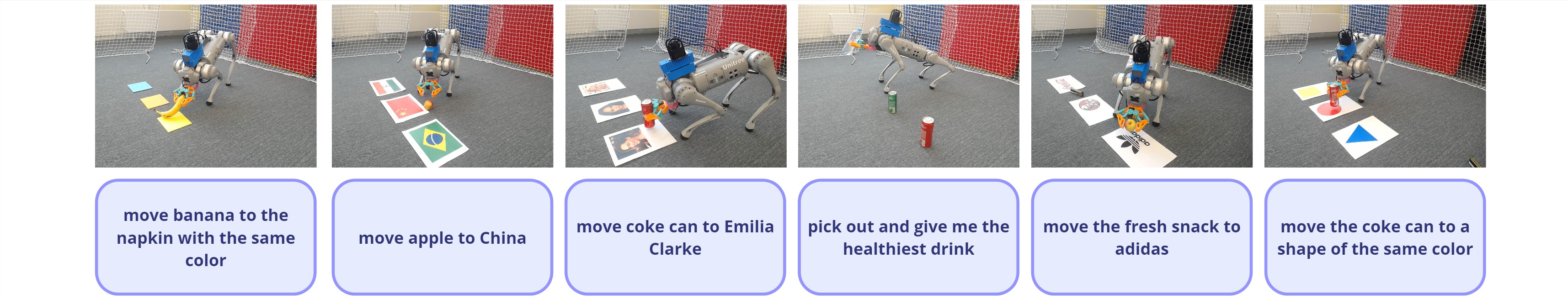}
  \caption{System Adaptability in Real-World Reasoning, Symbol Understanding, and Human Recognition Scenarios.}
  \Description{The figure consists of six images. On each image robot dog is performing the task which requires cognitive ability. The task, which robot needs to complete, is described as a caption under the image. Further we will describe each image separately.
  \begin{enumerate}
      \item[1.] Image 1. The caption under the image says: 'move banana to the napkin with the same color'. In the image robot dog is standing in front of three napkins of different colors, laying on the ground: blue napkin located far top, orange is in the middle, and yellow is close down. The robot dog is holding a banana with the gripper. The robot is near the yellow napkin and tilts down for placing a banana on it.
      \item[2.] Image 2. The caption under the image says:'move apple to China'. In the image robot dog is standing in front of the printed flags of three countries: India, China, and Brazil. The robot is close to the China flag and tilts down before it. The apple lies near this flag.
      \item[3.] Image 3. The caption under the image says: 'move coke can to Emilia Clarke'.  In the image the three printed faces of the three actresses (Margot Robbie, Zo\''e Kravitz, and Emilia Clarke) lay on the ground. The robot dog is standing in front of the face of Emilia Clark and tilts down to place coke can near it.
      \item[4.] Image 4. The caption under the image says: 'pick out and give me the healthiest drink'. In the image the two drinks lay on the ground. The drinks are Coca-Cola and Sprite. The robot dog stands near it and holds water bottle, tilting up to give it to the user.
      \item[5.] Image 5.  The caption under the image says: 'move the fresh snack to adidas'. The three printed logos lie on the ground. These logos are Coca-Cola, KFC and Adidas. The robot dog is standing in front of Adidas logo and holds an apple with the gripper. It tilts down to place an apple near the logo.
      \item[6.] Image 6.  The caption under the image says: 'move the coke can to a shape of the same color'. On the ground there are three paper pieces with printed shapes of different color: yellow square, red circle, and blue triangle. The robot dog is standing in front of the red square image and tilts down to place a coke can on it.
  \end{enumerate}}
  \label{fig:teaser}
\end{teaserfigure}

\maketitle

\section{Introduction}

The global scientific community is focused on creating a universal AI robot, driven by Large Language Models (LLMs) like OpenAI's ChatGPT \cite{lib:openai2022introducing} and later open source models \cite{lib:touvron2023llama}, \cite{almazrouei2023falcon}, \cite{jiang2023mistral}. Efforts to apply LLMs in robotics faced challenges, notably in understanding and processing the external world. Attempts to convey the model's understanding of the world around it through text-only approaches \cite{ahn2022can}, \cite{lib:singh2023progprompt}, \cite{hao2023reasoning} struggled with ambiguities and the assumption of static objects unless interacted with.

Multi-modal transformer-based models like GPT-4 \cite{openai2023gpt4} and Gemini \cite{Pichai2023}, which can process images, opened up new possibilities for robotics \cite{driess2023palm}. They enable robots to comprehend their environment, enhancing their `Embodied Experience' \cite{mu2023embodiedgpt}. However, many robots still face mobility and range limitations. The development of universally capable robots is now focusing on biologically-inspired designs, such as humanoid robots by Tesla Inc. \cite{Tesla_bot} and robotic dogs by Boston Dynamics \cite{lib:bostondynamics2023robotschat}, \cite{lib:petkauskas2023chatgpt}. However, the latter's interaction with the environment is limited to understanding and movement, lacking universality.

Our work builds on these foundations, aiming to create a robot that combines the strengths of sophisticated spatial action execution through environmental image analysis and biological designs. We chose the Unitree Go1 robot dog as our platform, equipped with a custom gripper and RGB-D cameras for vision. This setup enables the robot to navigate and interact with its environment effectively. We also implement a speech processing model for auditory command interpretation. The robot's cognition is developed using an `Inner Monologue' \cite{lib:huang2022inner} approach, facilitating task decomposition and memory retention. This framework is complemented by a communication system among multiple agent models, inspired by the `Autogen' \cite{wu2023autogen} concept, ensuring comprehensive and coherent robot behavior. In our system we use two transformer based models for behavior planning and for  environment analysis, which share information between each other in natural language.

\section{Related Works}

The field of robotic AI has witnessed notable advancements that closely align with our research. Key contributions include:

\textbf{Human Task Understanding in Manipulator Robots}:
Earlier works showcased robots capable of comprehending and executing human tasks \cite{driess2023palm}, \cite{brohan2022rt}, \cite{brohan2023rt}. These models, akin to LLM outputs, improved performance through an inner monologue approach, refining actions based on task feedback.

\textbf{Microsoft's Autogen Approach}:
Microsoft's Autogen project \cite{wu2023autogen} represents a shift from single-model inner monologues to multi-model conversations. Each model specializes in distinct areas, enhancing understanding and response diversity.

\textbf{Mistral 7B Model Integration}:
The Mistral 7B model excels in MMLU, commonsense reasoning, and world knowledge, outperforming Llama 2 13B in most evaluations \cite{jiang2023mistral}. Its superiority makes it an ideal candidate for generating robotic actions, considering environmental context and historical data.

\textbf{MiniGPT4 and Environmental Analysis}:
MiniGPT4 \cite{zhu2023minigpt}, \cite{chen2023minigptv2}, specializing in environmental analysis, coupled with the BLIP2 visual encoder, offers advanced image processing capabilities. This Visual Question Answering (VQA) model ensures universality in conveying nuances of environmental analysis tasks.

These advancements lay the foundation for our research, synthesizing elements into a unified, adaptable robotic system for seamless integration into human environments.

\section{Approach Overview}

\begin{figure}[h]
  \centering
  \includegraphics[width=\linewidth]{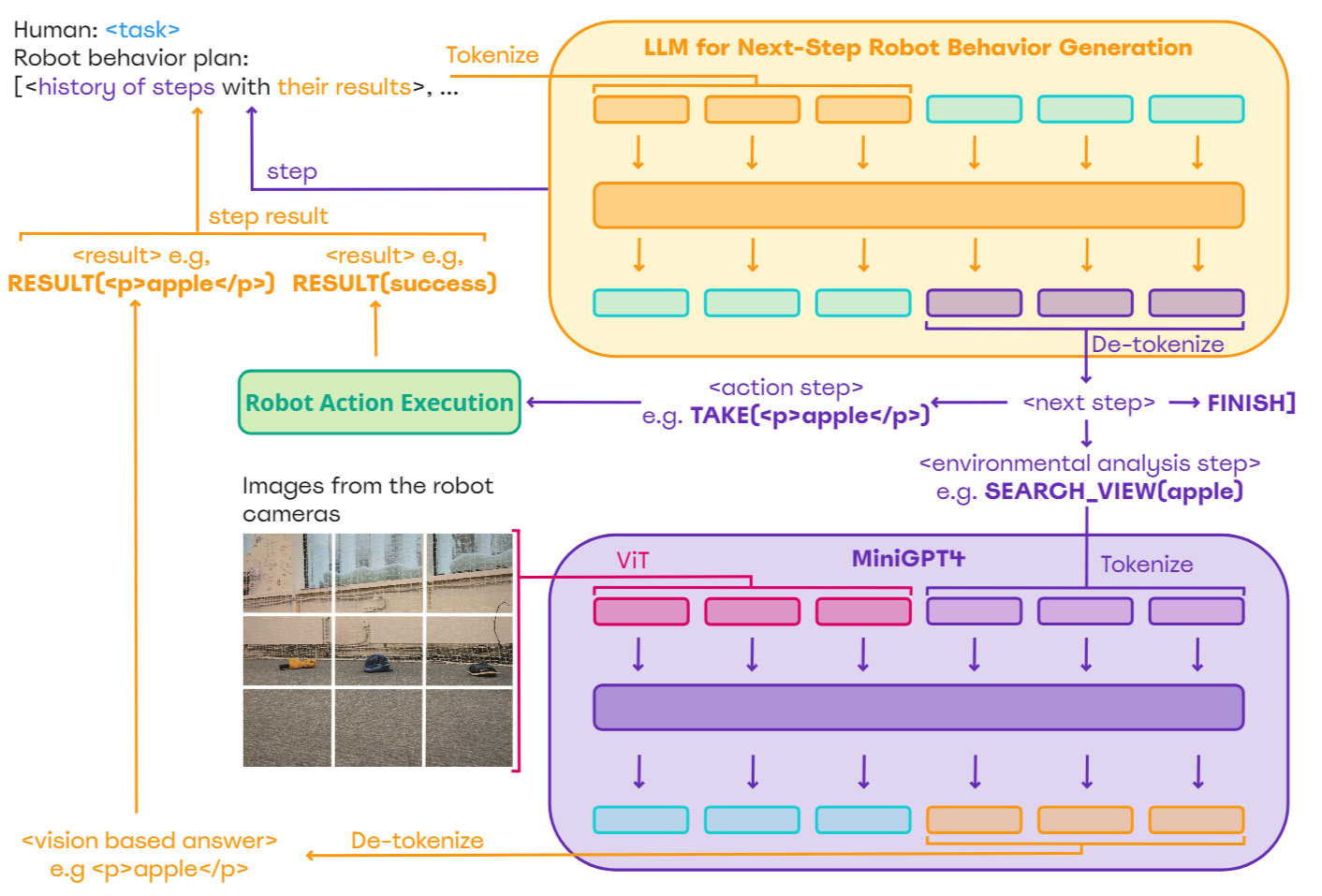}
  \caption{System Architecture of CognitiveDog}
  \Description{The system architecture describes the pipeline of developed approach. We will describe here the contents of the image step by step.
  \begin{enumerate}
      \item[1.] In the top left there are three lines of text. First line: 'Human: <task>'. The word '<task>' is colored blue. Second line: 'Robot behavior plan:'. The third line: '[history of steps with their resluts>, ... ]'. The words 'history of steps' are purple, the words 'their results' are orange. This part represent an initial prompt, which robot gets on each step of plan generation. The prompt is passed further to LLM.
      \item[2.] The large orange block in the top right of the image represents LLM for Next-Step Robot Behavior Generation (as the text at the top of that block says). The model tokenizes the prompt, keeps in memory the history of the prompt and then generates the next step based on the needed action. 
      \item[3.] Purple arrow comes down from LLM block to the words '<next step>'. Three arrows go from these words to the left, to the right and down. The arrow to the right leads to 'FINISH]' command. This step happens when the task is fully completed. The arrow to the left leads to '<action step>; e. g. TAKE(<p>apple</p>)'. This step happens when the robot dog needs to take action without communicating with visual language model. Next arrow to the left leads to block 'Robot Action Execution' and then sends result to the history of completed actions. The arrow down from the 'next step' leads to '<environmental analysis step>; e.g. SEARCH\_VIEW(apple)'. This step requires using MiniGPT4.
      \item[4.] MiniGPT4 tokenizes the command and uses ViT for image analysis depending on the task. It represented with purple block in the bottom right with little blocks of different colors (purple represents MiniGPT4, blue represents task given by a human, pink represents visual transformer and orange represents the results). After result is achieved, it is sent to the robot again. Near this block the example image from the robot cameras is situated. The image is split in 9 parts (it represents how the model analyzes this image). In the image there are three objects which lie on the floor: an umbrella, a winter hat and a slipper.
  \end{enumerate}}
  \label{fig:arch}
\end{figure}

In this section, we present a comprehensive overview of our methodology and the detailed process of developing a system based on this approach. The system's architecture is illustrated in Figure 2. Our primary focus is on the core model within this system, designed to assist the robot in selecting the subsequent steps towards accomplishing a given task. This includes a thorough examination of the methods used to compile a custom dataset of robot behaviors and the procedure for training a Large Language Model (LLM) on this dataset. Further, we delve into the application of a vision-language transformer model, employed without additional training, to provide the robot with a complete understanding of its surroundings. Finally, we will discuss the implementation aspects of executing various actions using the physical robot, specifically the Unitree Go1, a quadruped robot. This involves addressing the challenges and solutions in translating the planned actions into real-world movements and interactions.

\subsection{Step by Step Plan Generation}

The model for generating the next step in our work is pivotal as it shapes the robot's ability to execute tasks. It selects the optimal approach, analyzes the results and history of its actions, and adapts to changes. This behavior creation process is iterative, progressing step by step. At each iteration, the model receives a prompt composed of the operator's task and a history of actions taken, each followed by its outcome. Based on this data, the robot generates the next step toward achieving the goal. If the data indicates the user's task is completed, the model produces "FINISH" as the next step. 

The list of steps the model can generate is finite but fully encompasses the physical capabilities of the platform. Each potential step has its processing method, possible arguments, and returns a result post-execution, reflected in the step history. Steps include physical actions such as \textbf{"GO\_TO"}, \textbf{"TAKE"}, \textbf{"PUT\_IN"}, \textbf{"TILT"}, \textbf{"SIT"}, \textbf{"UP"}, \textbf{"TURN"}, \textbf{"SAY"}, \textbf{"FOLLOW"}, \textbf{"GIVE\_TO\_USER"}, \textbf{"GO\_USER"}, each representing a set of programs executed by the robot, usually resulting in success or failure.


More complex steps involve environmental analysis, such as \textbf{"DESCRIBE\_VIEW"}, \textbf{"QUESTION\_VIEW"}, and \textbf{"SEARCH\_VIEW"}. These functions engage another model with a visual encoder, which will be discussed in the next subsection. These steps bridge two large language models, with both input and output in free-form natural language. 


In crafting a universally adaptable robot for human environments, autonomy and independence from external sources are crucial. Therefore, we opted for models with around 7 billion parameters for their potential in autonomous operation. A technical framework for deploying 7B models on microcomputers with specifications akin to the Unitree Go1's onboard computer is detailed in an MIT study \cite{lib:mitref}. After evaluating existing 7B models, Mistral7B was selected as the base model. According to the model's creators \cite{jiang2023mistral}, Mistral7B outperforms the 13B version of LLaMa2 in the MMLU metric, equals LLaMa 34B in reasoning, and matches LLaMa2 13B in world knowledge. This high level of understanding in world context, text, and reasoning is vital for constructing the real-world robot plans.

In our research, we simplified the training of the model for environmental tasks by employing a specialized model for environmental analysis, which facilitated the collection of diverse datasets without being tied to the specific environments or objects. This approach ensures the model's effectiveness is not affected by changes in object appearances or environmental variations.

For fine-tuning, we created a dataset encompassing various robot operation scenarios. The fine-tuning process emphasizes teaching the model to construct steps iteratively. Instead of generating a dataset sample for every step, we employed a structure that includes a consistent system prompt, a user-defined task, and a comprehensive list of steps and outcomes for a scenario. The model, given this structure, predicts subsequent steps and outcomes, adding each new step and its outcome to the existing history. This strategy minimizes the number of required samples while maintaining their informational richness.

\subsection{Visual Information Analysis}

In this subsection, we explore how the MiniGPT4-v2 model \cite{chen2023minigptv2} analyzes and interacts with environments. Key capabilities include answering image-based questions, providing image part descriptions, and locating objects, each detailed below.

The images accompanying each query to this model are stitched together from three cameras positioned at the front and sides of the robot, offering a panoramic view. For image-based queries (\textbf{"QUESTION\_VIEW"}), the model uses the [vqa] tag for processing, yielding concise answers. In \textbf{"DESCRIBE\_VIEW"}, the argument supplemented by the prompt "describe shortly" for brief outputs. The object search feature (\textbf{"SEARCH\_VIEW"}) uses the [detection] tag and returns the object with its identifier and bounding box coordinates. These coordinates are transformed into 3D spatial data using Visual-SLAM, linked to the object identifier. The output includes the identifier, signaling the robot's interaction capability with the object. For objects with identical names, unique identifiers, such as \textbf{"SEARCH\_VIEW(apple), RESULT(<p>apple [1]</p> <p>apple [2]</p>)"}, are used.

\section{Evaluation}

Our experiments were aimed at assessing the model's ability to generate plans for diverse tasks of varying degrees of complexity. The main purpose of the first experiment is to evaluate the generalization property of the system over various unseen objects, backgrounds, and environments. The second experiment explores the ability to enable emergent capabilities, e.g. symbol understanding or human recognition. In the course of the experiments, we evaluated the system's capability to generate complex robot behavior in a dynamic environment, utilizing and adapting the experimental methodology outlined in the article RT-2 \cite{brohan2023rt} to suit our platform. 

The third experiment evaluates the ability of our system to construct plans for complex tasks based on the information obtained during completion of the previous steps. This feature comes from the mobility of the system, which allows to expand the knowledge of the environment and act in broadened space.

\subsection{Generalization}
For evaluation of model's generalization abilities, we collected the test dataset, consisting of instructions that were not used in training. We split the test data into three \textit{unseen} categories (objects, backgrounds, and environments) and additionally split into easy and hard cases, following this structure from RT-2. 

The evaluation consists of more than 130 instructions, and some of them are repeated for two or all three categories. We adapt the instructions from the list in RT-2. However, some objects were changed to similar ones, and we use the other environments and backgrounds.

\begin{figure}[h]
  \centering
  \includegraphics[width=\linewidth]{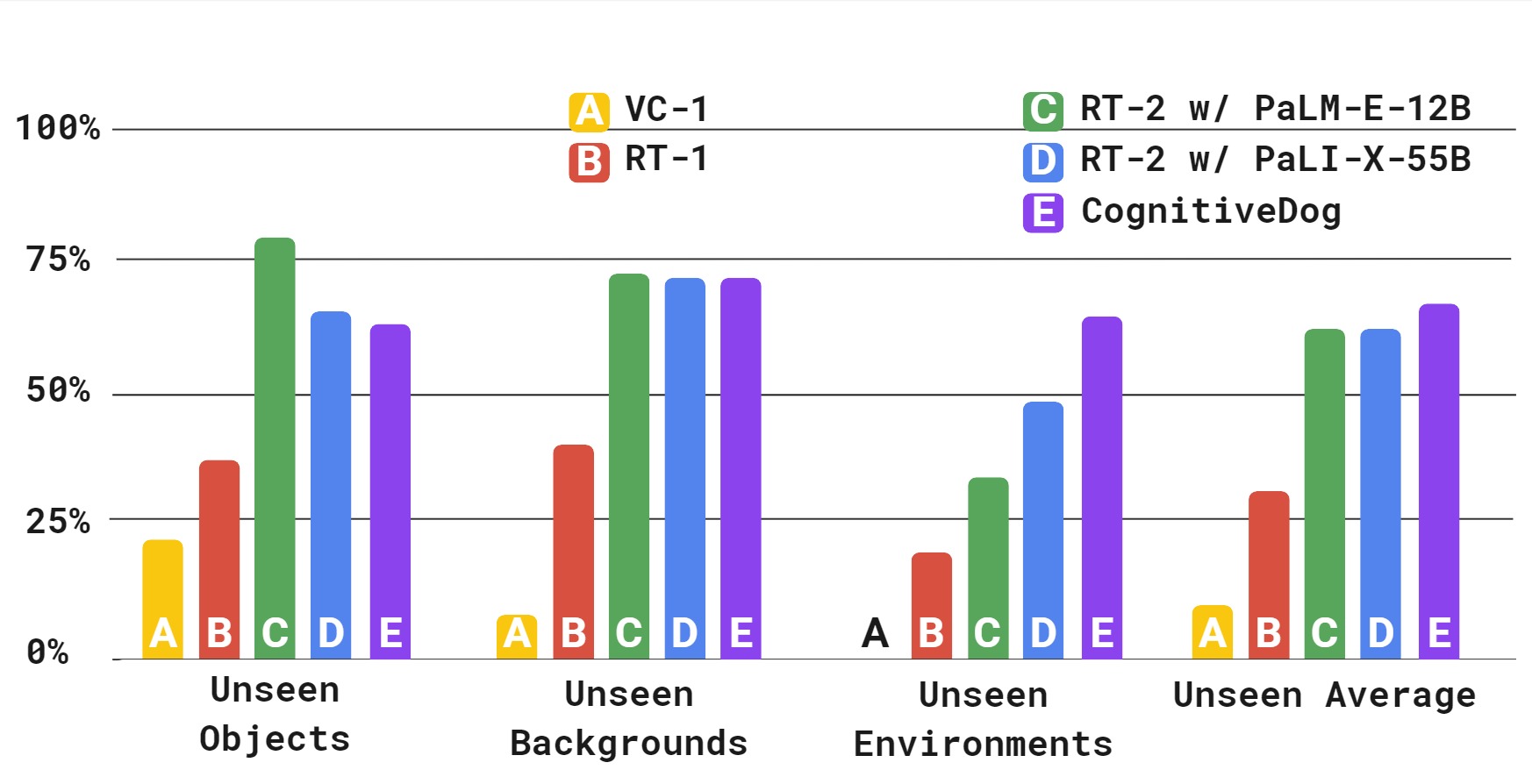}
  \caption{Evaluation of generalization abilities for various unseen categories.}
  \Description{ The results presented in the table are in percent for each category. Approaches names are the columns of the table, while category names are the rows. \begin{table}[h!]
    \centering
    \begin{tabular}{|c|c|c|c|c|c|}
        \hline
           Category result, \% / Approach name & VC-1 & RT-1 &  RT-2 w/ PaLM-E-12B & RT-2 w/ PaLI-X-55B & CognitiveDog \\     \hline
          Unseen Objects   & 22.0  & 37.0 & 66.0 & 75.5 & 65.5 \\     \hline
          Unseen Backgrounds & 8.0 & 40.0 & 72.0 & 73.0 & 73.9 \\     \hline
          Unseen Environments & 0 & 20.0 & 48.0 & 34.5 & 64.79 \\     \hline
          Unseen Average & 10.0 & 32.0 & 62.0 & 62.0 & 68.06 \\  \hline
    \end{tabular}
    \caption{Evaluation of generalization abilities for various unseen categories.}
    \label{tab:exp_1}
\end{table}}
  \label{fig:gen}
\end{figure}
The evaluation results are shown in Figure \ref{fig:gen}. Our system is designed and evaluated on the objects in real environment. Nonetheless, it still achieves results close to the laboratory conditions of RT-2 on objects category. Moreover, the results in backgrounds category are similar to both RT-2 model, and for the environments category our model outperforms all models. The observed results can be attributed to the high adaptability of MiniGPT4-v2 to different image input, as this model was trained to serve as a single interface optimized to efficiently perform a variety of visual-language tasks.
The average result for all three categories shows that our approach performs 3\% better than the largest RT-2 model, despite the fact that the total number of parameters of our models is almost four times less compared to RT-2 with PaLI-X. 



\subsection{Emergent Capabilities}
For the second experiment we also derive the split categories of emergent capabilities from RT-2 work: symbol understanding, reasoning and human recognition. We decided to expand the list of instructions for each category to test variability of our approach.

\begin{figure}[h]
  \centering
  \includegraphics[width=\linewidth]{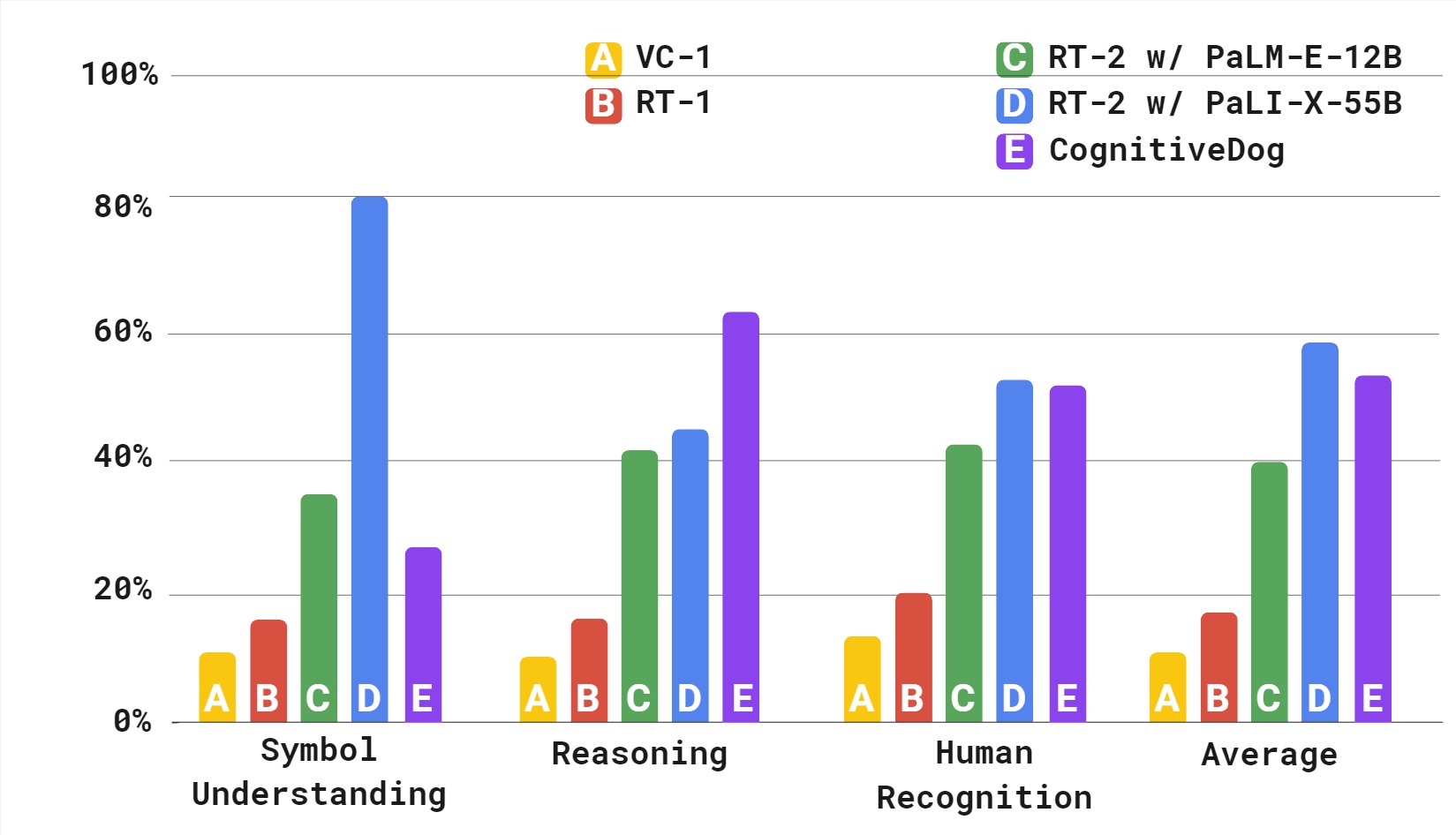}
  \caption{Performance comparison for three categories of emergent capabilities.}
  \Description{The image represents results of experiment for three categories of emergent capabilities. 
  \begin{table}[h!]
    \centering
    \begin{tabular}{|c|c|c|c|c|c|}
    \hline
           Category result / Approach name & VC-1 & RT-1 &  RT-2 w/ PaLM-E-12B & RT-2 w/ PaLI-X-55B & CognitiveDog \\     \hline
          Symbol understanding & 11 & 16 & 79 & 36 & 24  \\     \hline
          Reasoning & 10 & 16 & 46 & 42 & 62 \\     \hline
          Human Recognition & 11 & 17 & 55 & 42 & 54 \\     \hline
          Average & 10.3 & 16.3 & 59.0 & 38.6 & 55.0      \\     \hline
    \end{tabular}
    \caption{Performance comparison for three categories of emergent capabilities.}
    \label{tab:exp_2}
\end{table}}
  \label{fig:emerge}
\end{figure}

The results are presented in Figure \ref{fig:emerge}. We observe that our approach struggles in symbol understanding in comparison to RT-2 models, but still achieves better result than the baselines. This limitation is caused by MiniGPT4-v2 flaws and can be improved by additional VQA model training. We also note, that our approach shows the best result in reasoning category, outperforming both PaLM-E and PaLI-X RT-2 models. We attribute this result to the fact that our system design is inner monologue of two models, which is proven to significantly increase reasoning. Furthermore, Mistral7B achieves better results on all reasoning tasks \cite{jiang2023mistral}.


On average our approach is comparable with RT-2 models by  success rate and performs 16\% better than PaLM-E and only 4\% worse than PaLI-X. Importantly, we have achieved the best result in reasoning category, outperforming second best RT-2 with PaLI-X by 16\%, which shows the improved ability for plan generation.

\subsection{Complex Tasks Demonstration}
In this experiment we assess the system's ability to generate and execute plans for complex tasks, which require more sophisticated reasoning behavior. We do not fine-tune model for this experiment in particular. All the test complex tasks are unseen tasks for the system, and to complete the task it is usually required to get information from intermediate step of the plan and then act accordingly to the newly received data. For the experiment we have prepared the test list of human-written samples, containing complex tasks, e.g. "look at the picture, and then find and bring object suitable for that activity".

\begin{figure}[h]
  \centering
  \includegraphics[width=\linewidth]{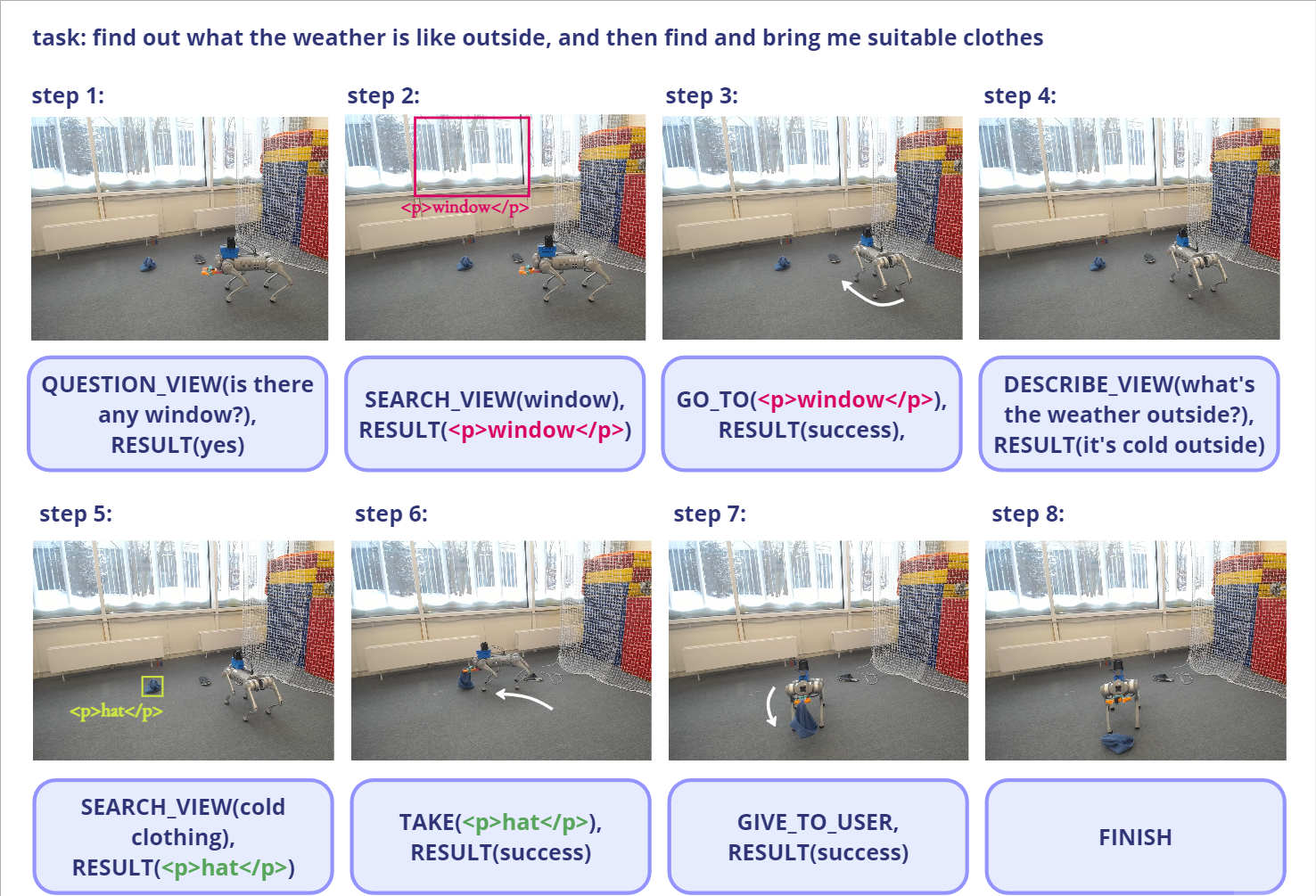}
  \caption{Step-by-step complex task completion by quadruped robot.}
  \Description{Image consists of 8 frames, each representing the step execution of the complex task. The text on top of the image is the task:'find out what the weather is like outside, and then find and bring me suitable clothes'. Each frame has a caption describing the action and its result for each step. Further we will describe every step separately.
  \begin{enumerate}
      \item[1.] Frame 1. The caption under the image says: 'QUESTION\_VIEW(is there any window), RESULT(yes)'. In the image robot dog is staying in the room with big window. The winter hat and the slipper lay on the ground.
      \item[2.] Frame 2. The caption under the image says:'SEARCH\_VIEW(window), RESULT(<p>window</p>)'. The image is the same as for previous step with the bounding box of detected window added.
      \item[3.] Frame 3. The caption under the image says:'GO\_TO(window), RESULT(success)'. In the image the robot dog has turned to the detected window.
      \item[4.] Frame 4. The caption under the image says:'DESCRIBE\_VIEW(what's the weather outside?), RESULT(it's cold outside)'. The image is the same as in the previous step.
      \item[5.] Frame 5. The caption under the image says:'SEARCH\_VIEW(cold clothing), RESULT(<p>hat</p>)'. In the image the bounding box of detected hat is shown. 
      \item[6.] Frame 6. The caption under the image says:'TAKE(<p>hat</p>), RESULT(success)'. In the image the robot dog takes the winter hat.
      \item[7.] Frame 7. The caption under the image says:'GIVE\_TO\_USER, RESULT(SUCCESS). In the image robot dog has turned to the camera and came closer to the user behind it.
      \item[8.] Frame 8. The caption under the image says:'FINISH'. The robot dog has put the hat on the floor in front of the user.
  \end{enumerate}}
  \label{fig:weather}
\end{figure}

We qualitatively observe that our system is able to both generate and execute plans for such complex tasks. We assume that such result can be explained by inner monologue architecture, which allows more diverse and comprehensive planning. The step-by-step completion of example complex task by our quadruped robot can be seen in Figure \ref{fig:weather}.

This experiment in particular shows that the mobility of the robot, the inner monologue architecture allows our system to operate in non-laboratory, open environments. Combined with natural language instructions, our approach allows natural and broad human-robot interaction.

\section{Conclusion}

In this study, we have introduced a groundbreaking application of inner monologue combined with several transformer-based models to enhance the capabilities of a quadruped robot. This novel approach enables the robot to generate and execute plans for a diverse range of real-world tasks. 
Our experimental evaluation has demonstrated the robot's high adaptability to varying environmental parameters. Notably, the robot's performance in Human Recognition and Reasoning tasks exceeded that of RT-2 with PaLM-E by 12\% and 20\% respectively, and approached the levels of RT-2 with PaLI-X, despite the latter having approximately four times more weights than the collective models in our system. In terms of reasoning ability, our system significantly outperformed the larger RT-2 model by 16\%. This superiority is attributed to the effective combination of the high reasoning capacity of the base model and the inner monologue approach. Furthermore, the integration of this advanced system into a robust quadruped robot platform, featuring an almost unlimited working area, brings us closer to realizing a universal robot. Such a robot would be capable of providing natural and comprehensive human-robot interaction, marking a significant advancement in the field of robotics and artificial intelligence.


\bibliographystyle{ACM-Reference-Format}
\balance
\bibliography{lib}

\appendix









\end{document}